\documentclass[11pt]{article} 

\usepackage{rldmsubmit,palatino}
\usepackage{graphicx}
\usepackage[square,numbers]{natbib}
\usepackage{xcolor}
\definecolor{mydarkblue}{rgb}{0,0.08,0.45}
\usepackage[colorlinks = true,
            linkcolor = mydarkblue,
            urlcolor  = mydarkblue,
            citecolor = mydarkblue,
            anchorcolor = mydarkblue]{hyperref}
\usepackage{glossaries}
\usepackage{amsmath}
\usepackage{amssymb}
\usepackage{bm}
\usepackage{algorithm}
\usepackage{algorithmic}
\newacronym{rl}{RL}{Reinforcement Learning}
\newacronym{drl}{DRL}{Deep Reinforcement Learning}
\newacronym{mdp}{MDP}{Markov Decision Process}
\newacronym{ppo}{PPO}{Proximal Policy Optimization}
\newacronym{sac}{SAC}{Soft Actor-Critic}
\newacronym{mlp}{MLP}{Multilayer Perceptron}
\newacronym{epvf}{EPVF}{Explicit Policy-conditioned Value Function}
\newacronym{unf}{UNF}{Universal Neural Functional}

\title{Massively Scaling Explicit Policy-conditioned Value Functions}

\author{
Nico~Bohlinger \\
Department of Computer Science\\
Technical University of Darmstadt, Germany\\
\texttt{nico.bohlinger@tu-darmstadt.de} \\
\And
Jan Peters \\
German Research Center for AI (DFKI) \\
Hessian.AI \\
Centre for Cognitive Science \\
Department of Computer Science \\
Technical University of Darmstadt, Germany \\
\texttt{jan.peters@tu-darmstadt.de} \\
}

\begin{document}

\maketitle

\begin{abstract}
We introduce a scaling strategy for Explicit Policy-Conditioned Value Functions (EPVFs) that significantly improves performance on challenging continuous-control tasks. EPVFs learn a value function $V(\theta)$ that is explicitly conditioned on the policy parameters, enabling direct gradient-based updates to the parameters of any policy. However, EPVFs at scale struggle with unrestricted parameter growth and efficient exploration in the policy parameter space.
To address these issues, we utilize massive parallelization with GPU-based simulators, big batch sizes, weight clipping and scaled peturbations.
Our results show that EPVFs can be scaled to solve complex tasks, such as a custom Ant environment, and can compete with state-of-the-art Deep Reinforcement Learning (DRL) baselines like Proximal Policy Optimization (PPO) and Soft Actor-Critic (SAC).
We further explore action-based policy parameter representations from previous work and specialized neural network architectures to efficiently handle weight-space features, which have not been used in the context of DRL before.
\end{abstract}


\acknowledgements{
This project was funded by National Science Centre, Poland  under the OPUS call in the Weave program UMO-2021/43/I/ST6/02711, and by the German Science Foundation (DFG) under grant number PE 2315/17-1.
}

\startmain

\section{Introduction}

The remarkable success of deep learning in recent years is closely linked to the concept of \emph{scaling}.
In particular, the scaling of neural networks, datasets, and computational resources has led to significant advances in various domains such as computer vision and natural language processing.
Hardware improvements for GPUs and TPUs and parallelization techniques like data and model parallelism have enabled the training of large-scale models on massive datasets.
The relationship between scaling and improved performance has been formalized into the concept of \emph{scaling laws}~\citep{kaplan2020,hoffmann2022}.

While the benefits of scaling are well-established in supervised learning, its application to \gls{drl} is less understood.
Naively increasing the size of policy and value function networks and using bigger batch sizes often leads to diminishing returns or even performance degradation~\citep{andrychowicz2021}.
However, recent work has shown that carefully designing the neural network architecture by incorporating normalization layers, like LayerNorm, BatchNorm or WeightNorm, and residual connections can help to scale up \gls{drl} algorithms and improve their performance with bigger networks~\citep{kumar2023,bhatt2024,nauman2024,lee2024}.
On-policy algorithms like \gls{ppo} have been shown to greatly benefit from bigger batch sizes, which can be collected efficiently using up to thousands of parallel environments~\citep{rudin2022}.

In this work, we investigate the scaling capabilities of different variations of \glspl{epvf} with the help of massively parallel environments.
While \glspl{epvf} have previously struggled on complex tasks, we show the importance of scaling and weight regularization for the training of \glspl{epvf} and compare different neural network architectures and training setups on a MuJoCo Ant and Cartpole environment.

\section{Explicit Policy-conditioned Value Functions}
In \gls{rl}, we consider a \gls{mdp} defined by a tuple $\mathcal{M} = (\mathcal{S}, \mathcal{A}, P, R, \gamma, \rho_0)$, where $\mathcal{S}$ is the state space, $\mathcal{A}$ is the action space, $P$ is the transition dynamics, $R$ is the reward function, $\gamma$ is the discount factor, and $\rho_0$ is the initial state distribution.
The goal of an \gls{rl} agent is to learn a policy $\pi_{\bm{\theta}}(a|s)$ that is parameterized by $\bm{\theta}$.
Rolling out the policy in the environment generates a trajectory $\tau = (s_0, a_0, r_0, \ldots)$, where $s_0$ is sampled from $\rho_0$ and $a_t \sim \pi_{\bm{\theta}}(\cdot|s_t)$.
The return $R_t$ is defined as the sum of discounted rewards $R_t = \sum_{k=0}^{T - t - 1} \gamma^k r_{t+k+1}$, where $T$ is the time horizon.
The policy is trained to maximize the expected return $J(\pi_{\bm{\theta}}) = \mathbb{E}_{\tau \sim \pi_{\bm{\theta}}, s_0 \sim \rho_0} \left[ R_0 \right]$.
The state-value function for the policy is defined as $V^{\pi_{\bm{\theta}}}(s) = \mathbb{E}_{\pi_{\bm{\theta}}} \left[ R_t | s_t = s \right]$, with which we can re-formulate the policies objective as
\begin{equation}
    J(\pi_{\bm{\theta}}) = \int_{\mathcal{S}} \rho_0(s) V^{\pi_{\bm{\theta}}}(s) \, ds.
\end{equation}

Many \gls{rl} algorithms use the action-value function $Q^{\pi_{\bm{\theta}}}(s, a) = \mathbb{E}_{\pi_{\bm{\theta}}} \left[ R_t | s_t = s, a_t = a \right]$ to decompose the state-value function in the policy objective into $V^{\pi_{\bm{\theta}}}(s) = \int_{\mathcal{A}} \pi_{\bm{\theta}}(a|s) Q^{\pi_{\bm{\theta}}}(s, a) \, da$ to get the gradient w.r.t. $\bm{\theta}$ to optimize the policy.
\glspl{epvf} take a different approach by learning a value function that is explicitly conditioned on policy parameters or some other differentiable representation of the policy $V(\bm{\theta})$, to directly optimize the gradient of the policy objective~\citep{harb2020,faccio2021,faccio2022}:
\begin{equation}
    \nabla_{\bm{\theta}} J(\pi_{\bm{\theta}}) = \int_{\mathcal{S}} \rho_0(s) \nabla_{\bm{\theta}} V(s,\bm{\theta}) \, ds = \mathbb{E}_{s \sim \rho_0} \left[ \nabla_{\bm{\theta}} V(s,\bm{\theta}) \right] = \nabla_{\bm{\theta}} V(\bm{\theta}).
\end{equation}

The value function $V(\bm{\theta})$ is learned using a replay buffer of policy parameters and returns, which can be collected by any policy, and updated using stochastic gradient descent.
Because the value function predicts the performance of a policy from start to finish and we consider the episodic setting, the target is the undiscounted return.
The policy is then updated by following the gradient of the value function with respect to the policy parameters.
The resulting algorithm is presented in Algorithm~\ref{alg:epvf}.

\begin{algorithm}[H]
    \caption{Actor-Critic with Explicit Policy-conditioned Value Function}
    \label{alg:epvf}
    \begin{algorithmic}
        \STATE Input: Initial policy parameters $\bm{\theta}$, initial value function parameters $\bm{\phi}$, replay buffer $D$
        \FOR{$I$ steps}
            \STATE Choose $M$ policy parameters or representations $\{\bm{\theta}_1, \ldots, \bm{\theta}_{M}\}$
            \STATE Rollout and compute undiscounted return $R_m$ for each $\bm{\theta}_m$
            \STATE Store $\{(R_1, \bm{\theta}_1), \ldots, (R_{M}, \bm{\theta}_{M})\}$ in $D$
            \FOR{$K$ steps}
                \STATE Sample batch $B = \{(R, \bm{\theta})_1, \ldots, (R, \bm{\theta})_{N}\}$ from $D$
                \STATE Update $\bm{\phi}$ with gradient descent: $\nabla_{\bm{\phi}} \frac{1}{N} \sum_{n=1}^N [R_n - V_{\bm{\phi}}(\bm{\theta}_n)]^2$
            \ENDFOR
            \FOR{$L$ steps}
                \STATE Update $\bm{\theta}$ with gradient descent: $\nabla_{\bm{\theta}} \frac{1}{N} \sum_{n=1}^N -V_{\bm{\phi}}(\bm{\theta})$
            \ENDFOR
        \ENDFOR
    \end{algorithmic}
\end{algorithm}

\glspl{epvf} have the key advantage to learn a value function that can potentially reason about all possible policies instead of just the current policy and directly optimize the policy through the value function network.
This enables completely off-policy or offline learning with any kind of policy data.
Data can be collected by policies that might be optimized for different objectives as long as the associated return for rolling out the policy is known.
This can be especially useful in multi-task settings or settings where exploration with different strategies is needed.

When using the concatenation of the raw policy parameters as the input to the value function, the size of the value function network can quickly blow up with the number of policy parameters.
This can be mitigated by using another representation of the policy, like the concatenation of actions given a set of probing states~\citep{harb2020,faccio2022}.
Furthermore, specialized neural network architectures that are designed to handle weight-space features efficiently, by utilizing the symmetry properties of neural network weights~\citep{navon2023,zhou2024a,zhou2024b}, are another interesting direction to prevent the network from blowing up in size.
These architectures have not been used in the context of \gls{rl} or even continual online learning before.
We investigate both alternatives and more in the following experiments.

\section{Experiments}
\begin{figure}[H]
    \centering
    \includegraphics[width=0.6\textwidth]{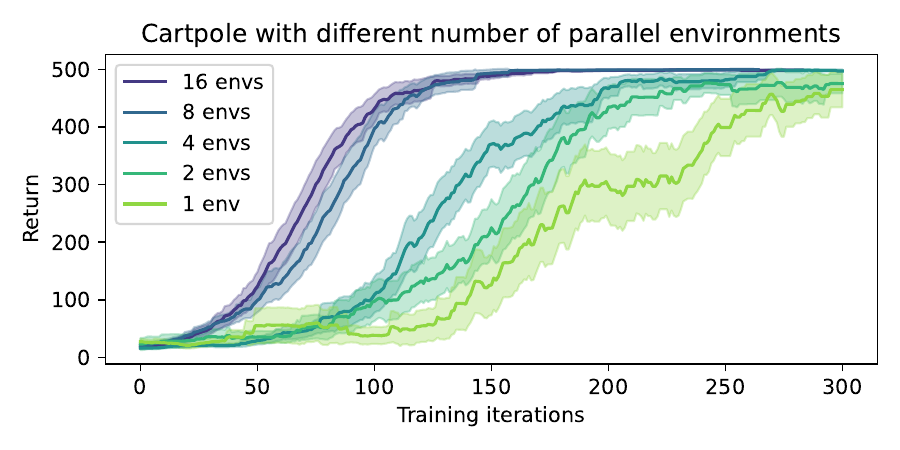}
    \caption{
        Performance on Cartpole with different numbers of parallel environments.
        The return is the average undiscounted return achieved with the perturbed policy parameters used during data collection.
    }
    \label{fig:cartpole}
\end{figure}

First, we evaluate the performance of \glspl{epvf} on the Gymnasium Cartpole environment~\citep{towers2024}.
From previous work, we know that \glspl{epvf} can fully solve simple tasks like Cartpole without additional scaling, i.e., using a small batch size of $16$, a replay buffer of size $1e5$, a two layer \gls{mlp} with $64$ neurons for the deterministic policy, and only a single environment~\citep{faccio2021}.
For choosing the policy parameters during rollout, we follow \citet{faccio2021} and simply use the current best policy parameters and perturb them with Gaussian noise $\mathcal{N}(\mu = 0, \sigma = 1.0)$.
We compare the single environment setting with using up to $16$ parallel environments, where the policy parameters for every environment are perturbed with a different sample of Gaussian noise, i.e. we use as many different policies as environments.
Figure~\ref{fig:cartpole} shows that scaling the number of environments, and therefore the number of different policies, improves convergence speed, while the perturbed policies in all settings eventually reach the same maximum return of $500$.
As the Cartpole task is not particularly challenging, we observe diminishing returns with more than $8$ environments.

Next, we evaluate the performance of \glspl{epvf} when scaling to massively parallel environments on a custom Ant environment.
For the physics simulation, we us MJX, which is a highly parallelizable GPU-based version of MuJoCo~\citep{todorov2012} based on JAX~\citep{jax2018}.
Our Ant environment is a continuous control task with a 34-dimensional state space and a 8-dimensional action space.
In this task, the Ant's objective is to walk forward at a target velocity of 2 m/s.
The time horizon is set to $1000$ steps and the reward is calculated as $r = \exp(- |v_{xy} - c_{xy}|^2 / 0.25)$, where $v_{xy}$ is the linear velocity of the Ant and $c_{xy}$ is the target velocity $(2, 0)$.
This results in a maximum possible return of $1000$.
We setup the learning environment and algorithm in the \gls{drl} framework RL-X~\citep{bohlinger2023}. 
We can jit-compile the full training loop, enabling up to $4096$ parallel environments on a single RTX 3090 GPU.
This setup allows for extremely fast data collection and throughput of up to $3$ million environment steps per second.

\begin{figure}[H]
    \centering
    \includegraphics[width=0.49\textwidth]{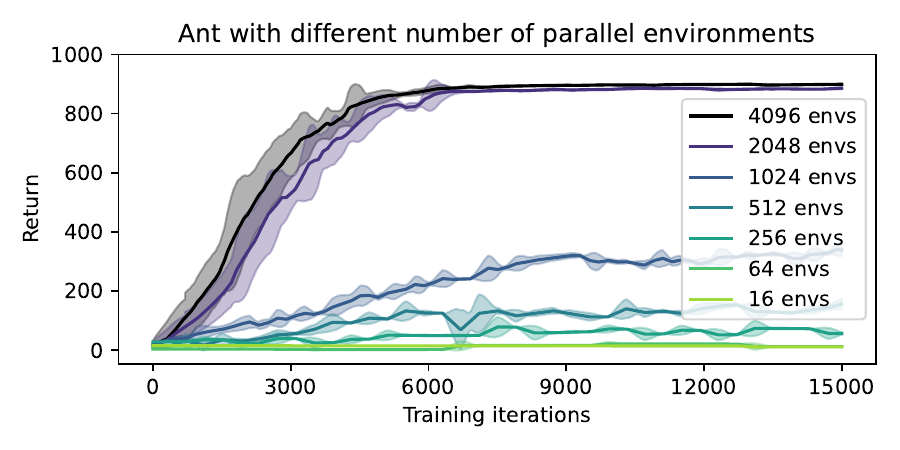}
    \includegraphics[width=0.49\textwidth]{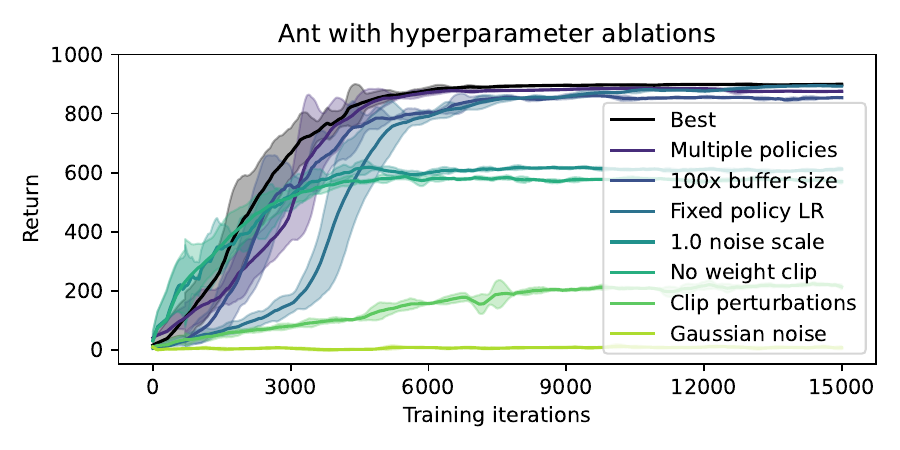}
    \includegraphics[width=0.49\textwidth]{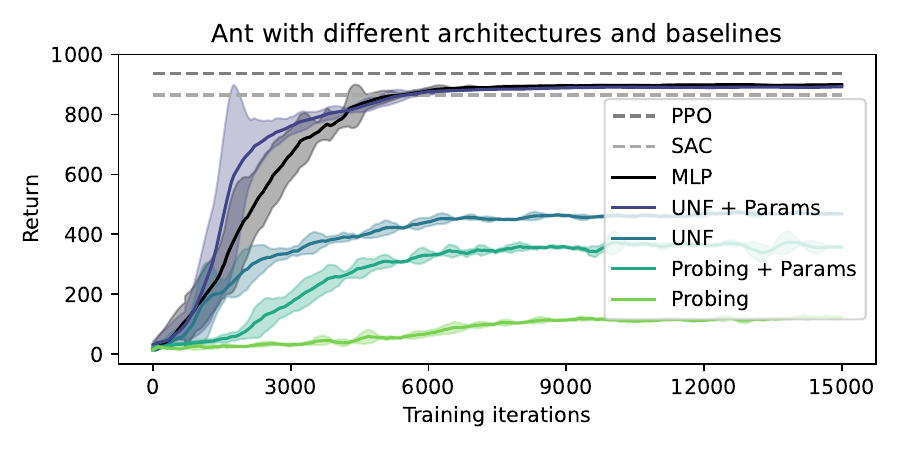}
    \caption{
        \emph{Top left} -- Performance on Ant with different numbers of parallel environments, which equal the batch size.
        \emph{Top right} -- Ablation on the different algorithmic changes introduced to scale \glspl{epvf}.
        Every ablation uses all the same parameters as the best setup but with one change.
        Multiple policies: Update a set of 4096 differently initialized policy parameters instead of just one.
        100x buffer size: Increase the replay buffer size to 409600 to sample older data as well.
        Fixed policy LR: Use a fixed learning rate of $1e-5$ for the policy instead of the learning rate schedule.
        1.0 noise scale: Use a uniform noise scale of $1.0$ for perturbing the policy parameters instead of $0.3$.
        No weight clip: Do not clip the policy parameters.
        Clip perturbations: Directly clip the perturbations of the policy parameters to $(-0.3, 0.3)$ instead of only after the gradient steps.
        Gaussian noise: Use $\mathcal{N}(\mu = 0, \sigma = 1.0)$ instead of uniform noise for perturbing the policy parameters.
        \emph{Bottom} -- Performance of the action-based policy parameter representations (Probing) and specialized weight-space architectures (\gls{unf}). Additionally, the performance of the \gls{ppo} and \gls{sac} baselines is shown as dashed lines.
    }
    \label{fig:ant}
\end{figure}

To successfully train \glspl{epvf} on the Ant environment, the training setup is significantly scaled up. We increase the number of environments to $4096$, the batch size to $4096$ and also the replay buffer size to $4096$.
As the replay buffer is now just as big as the batch size, the value function is trained on only fresh data and the big batch size ensures good gradient estimates through averaging on diverse data.
Furthermore, we perturb the policy parameters with uniform noise that is $(-0.3 \omega, 0.3 \omega)$ for each parameter $\omega$ instead of Gaussian noise.
This change allows the added noise to be automatically scaled to the magnitude of every policy parameter.
Additionally, we introduce weight clipping for the policy parameters to be in the range $(-0.1, 0.1)$, which helps to stabilize the training and prevent the policy parameters from constantly growing in magnitude.
Alternatively, weight regularization with strong weight decay can also be used but the weight decay coefficient is harder to tune.
Finally we add an exponential learning rate schedule for the policy from $1e-3$ to $1e-7$ over the course of training.
The top right of Figure~\ref{fig:ant} shows the ablations on the mentioned algorithmic changes.

The top left of Figure~\ref{fig:ant} highlights the scaling capabilities of \glspl{epvf} on the Ant environment, where the number of environments equals the batch size.
The results show that increasing the batch size directly improves the policy performance.
An environment and batch size of at least $2048$ is necessary to achieve strong performance and to solve the task, while setups with less environments struggle to reach even a quarter of the maximum return.
\glspl{epvf} thrive in settings with many environments and large batch sizes, as the value function trained for the policy parameter space is inherently more difficult to learn but can give good gradient estimates to the policy when the policy parameter space is well explored and well restricted with weight clipping.

Finally, we investigate action-based policy parameter representations and specialized neural network architectures for the value function and also compare the performance of \glspl{epvf} to \gls{ppo} and \acrfull{sac}~\citep{haarnoja2018}.
For the action-based policy parameter representation, we follow prior work~\citep{harb2020,faccio2022} and use the concatenation of actions given a set of 200 learnable probing states as the input to the value function.
Furthermore, we test a variation where the resulting action representation is concatenated with the raw policy parameters to see if both representations can be combined.
Figure~\ref{fig:ant} shows that the action-based representation is not reaching the same performance as just using the raw policy parameters.
The combination of both improves upon only using the actions but still does not reach good performance.\\
For the specialized weight-space neural network architecture, we use the recently proposed \acrfull{unf}~\citep{zhou2024b} for the value function.
Figure~\ref{fig:ant} shows that the features from the \gls{unf} architecture in concatenation with the raw policy parameters are able to reach the same performance as the default \gls{mlp} architecture, while even learning slightly faster.
When using \gls{unf} features alone, the learned policy is not able to reach the same end performance.

\section{Conclusion}
We have shown that \glspl{epvf} can be scaled to solve complex continuous control tasks and compete with state-of-the-art \gls{drl} baselines with the help of massively scaling up the number of parallel environments and the batch size.
Key to stability is the use of weight clipping, to restrict the policy parameter space and prevent the parameters from constantly growing, and the use of uniform noise scaled to the magnitude of the parameters for exploring the policy parameter space efficiently.
Further investigation on weight-space features from specialized architectures like \glspl{unf} might enable even better scaling capabilities for \glspl{epvf} in the future.

\bibliographystyle{corlabbrvnat}
\bibliography{bibliography}

\end{document}